# IDAN: Image Difference Attention Network for Change Detection

Hongkun Liu, Zican Hu, Qichen Ding, and Xueyun Chen *

***Abstract*—Remote sensing image change detection is of great importance in disaster assessment and urban planning. The mainstream method is to use encoder-decoder models to detect the change region of two input images. Since the change content of remote sensing images has the characteristics of wide scale range and variety, it is necessary to improve the detection accuracy of the network by increasing the attention mechanism, which commonly includes: Squeeze-and-Excitation block, Non-local and Convolutional Block Attention Module, among others. These methods consider the importance of different location features between channels or within channels, but fail to perceive the differences between input images. In this paper, we propose a novel image difference attention network (IDAN). In the image preprocessing stage, we use a pre-training model to extract the feature differences between two input images to obtain the feature difference map (FD-map), and Canny for edge detection to obtain the edge difference map (ED-map). In the image feature extracting stage, the FD-map and ED-map are input to the feature difference attention module and edge compensation module, respectively, to optimize the features extracted by IDAN. Finally, the change detection result is obtained through the feature difference operation. IDAN comprehensively considers the differences in regional and edge features of images and thus optimizes the extracted image features. The experimental results demonstrate that the F1-score of IDAN improves 1.62% and 1.98% compared to the baseline model on WHU dataset and LEVIR-CD dataset, respectively. The code we use will be available for download.**

***Index Terms*—Change detection; feature difference map; edge difference map; feature difference attention module; edge compensation module**

## I. INTRODUCTION

REMOTE sensing image change detection is of great significance in natural disaster assessment and land resource management, but the images collected at different times often have different styles, which brings challenges to high-precision detection. Image difference method [1] is a traditional method to detect the changing areas between images. The objects of difference include pixels, colors, textures, etc. Fangrong Zhou et al. [2] used the inter-frame image difference method to extract the features of the natural disaster area in the image. Fan Gao et al. [3] combined color features and texture features to improve the detection performance of the image difference method for changing regions. This method can easily detect the location of the changing area, but because the algorithm is relatively simple, the detection results are often rough and susceptible to interference from background noise. In order to achieve higher-precision in remote sensing image change detection, more researchers use deep convolutional features as the difference object: Lu Xu et al. [4] proposed a change detection method based on an improved U-net [5] to reduce the salt-and-pepper phenomenon and false detections in traditional methods. Compared with traditional methods, deep learning methods have a higher number of parameters and a complex model structure, so they have better experimental results.

Adding an attention mechanism to deep learning can improve the accuracy and efficiency of image change detection. The existing attention mechanisms mainly include three categories: (1) Channel attention mechanisms: Jie Hu et al. [6] used a fully connected layer to generate a weight for each channel to represent the importance of that channel in the next stage, and multiplied the weight with the corresponding channel as the output of this stage. (2) Spatial attention mechanism: Xiaolong Wang et al. [7] used a non-local module to make the neural network convolutional layer realize the global perception of the feature map, in order to find which area of the target that needs attention in the entire space. (3) Hybrid attention mechanism: Sanghyun Woo et al. [8] used the convolutional block attention module (CBAM), which can guide the neural network to learn what needs to be paid attention to in the channel and space dimensions at the same time. Qian Shi et al. [9] applied CBAM to the task of remote sensing image change detection and proved the effectiveness of the improvement. Xueli Peng et al. [10] added an attention mechanism to U-net++ [11] to optimize the performance of the model on seasonal change detection tasks. These attention mechanisms are optimized for the feature of a single image, without considering the feature difference and edge difference between the images, so it is difficult to perform feature optimization in the task of remote sensing image change detection.

In view of the limitations of the aforementioned methods, this article proposes image difference attention Network (IDAN). The backbone of IDAN is responsible for extracting image features, and then splitting and subtracting the channels of image features to obtain the detection result of the changed area. In order to perceive the changes in the input images, we input two images into a pre-training model to extract features and perform the difference operation to obtain the feature

This work was supported by the National Natural Science Foundation of China under Grant 62061002. (Corresponding author: Xueyun Chen.)

The authors are with the School of Electrical Engineering, Guangxi University, Nanning 530004, China (e-mail: 3039886255@qq.com)

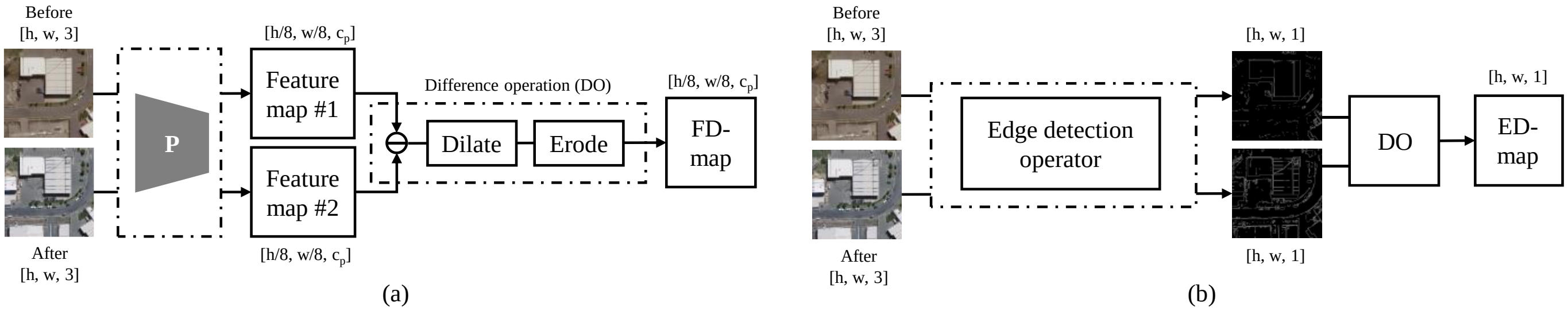

Fig. 1. The preparation process of the difference map. The shape of the corresponding data is shown in parentheses and $c_p$ represents the number of output channels of P. (a) The production process of FD-map. P means pre-training model. The specific process of the difference operation is shown in the dashed box. (b) The process of making ED-map. The difference operation (DO) is as shown in Fig. 1(a).

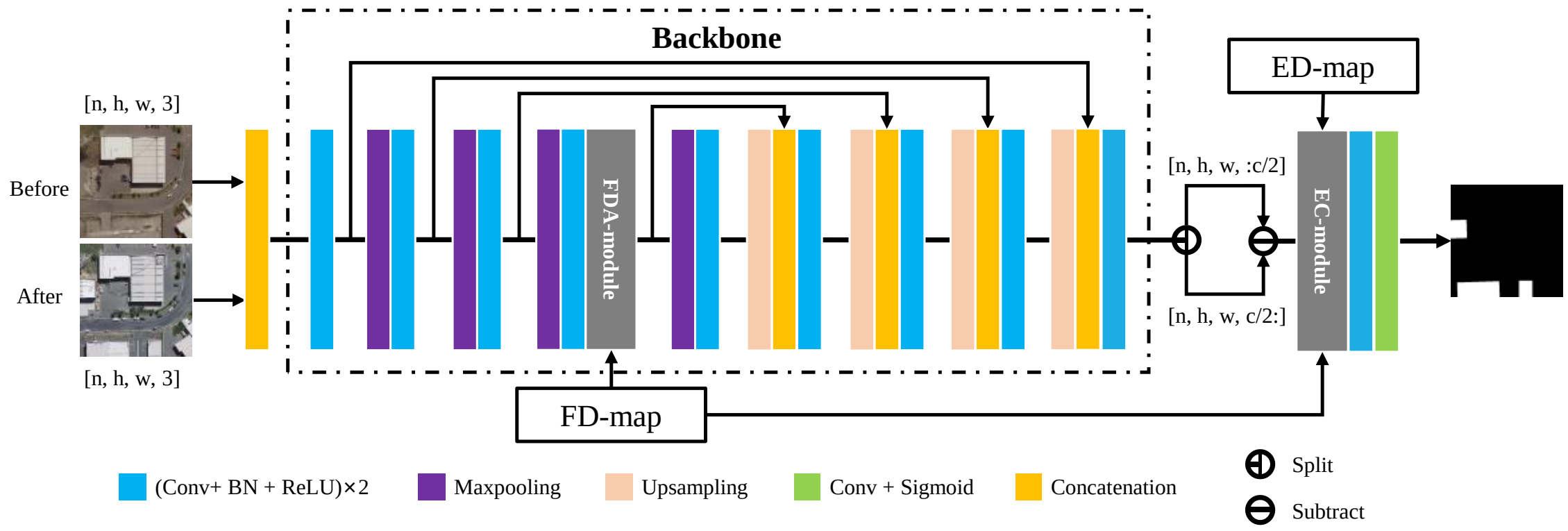

Fig. 2. IDAN network structure. The input of the network is two images, and the output is the corresponding change detection result. The area in the dashed box in the figure is the backbone.

difference map (FD-map), and then input the FD-map into IDAN through the feature difference attention module (FDA-module) to optimize the features. In order to improve the edge missing phenomenon in the feature extraction process of the model, we perform edge detection and difference operation on two images to obtain the edge difference map (ED-map), and then fuse the ED-map with the feature map in IDAN through the edge compensation module (EC-module). Experiments demonstrate that the proposed attention mechanism optimizes the model better than the aforementioned existing methods, and the F1-score of IDAN experimental results improves by 1.62% and 1.98% compared to the baseline model on two publicly available datasets.

The main contributions of this paper are as follows:

(1) IDAN is proposed for change detection in remote sensing images, and show better results than traditional methods.

(2) FD-map and FDA-module are designed to optimize the feature map within IDAN and improve the detection accuracy.

(3) By using ED-map, EC-module is designed to compensate the edges of features within IDAN to enhance the performance of model detection.

## II. OUR APPROACH

In this section we will introduce: (1) preparation of the difference map; (2) structure of IDAN; (3) two feature optimization modules in IDAN; (4) the feature difference method and the loss function of IDAN.

### A. *Preparation of difference map*

The preparation process of the FD-map is shown in Fig. 1(a). A pre-training model from the ImageNet [12] is used to extract the features of the two images and perform the difference operation. The function of FD-map is to use the image differrence method based on deep convolution feature to pre-extract the change area of the image, and to weight the corresponding position of the feature map in IDAN. The preparation process of ED-map is shown in Fig. 1(b). The edge operator is used to detect the edges of the two images respectively, and then the detection results are subjected to the difference operation. The function of ED-map is to use traditional image processing methods to pre-extract the edge information of the image in order to compensate for the lack of feature edges in the process of extracting image features by IDAN.

### B. *Network structure*

The structure of IDAN is shown in Fig. 2. The input is two images and the output is the detection result of the image change area. The two images are sent to the backbone for feature extraction after channels concatenated, and the parameters of the convolutional layer in the backbone are consistent with U-net [5]. After the backbone completes feature extraction, each channel of the feature map contains the information of the two input images. In order to directly obtain the feature difference between the input images, the feature map is split equally into two parts on the channel which then are subtracted. Finally, the convolutional layer is used to reduce the number of feature channels and get the change detection result. In IDAN, the feature map is optimized through the feature difference attention module (FDA-module), and the edge of the feature map is compensated through the edge compensation module (EC-module).

### C. *Feature optimization module*

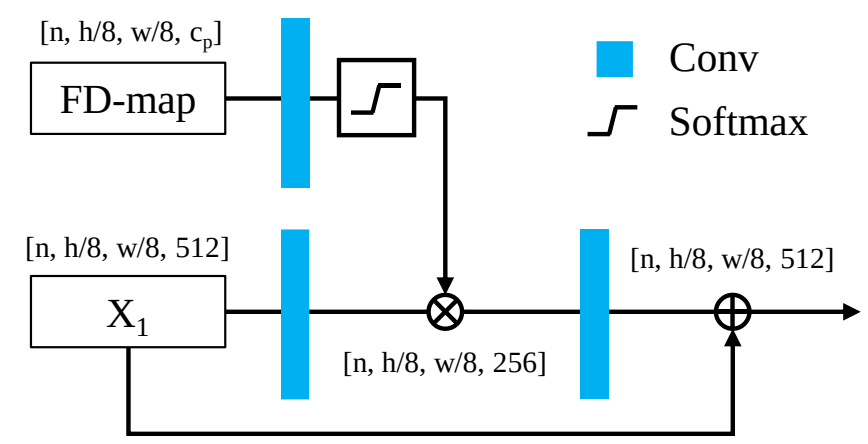

Fig. 3. Feature difference attention module. $X_1$ represents the feature map that needs to be optimized in IDAN.

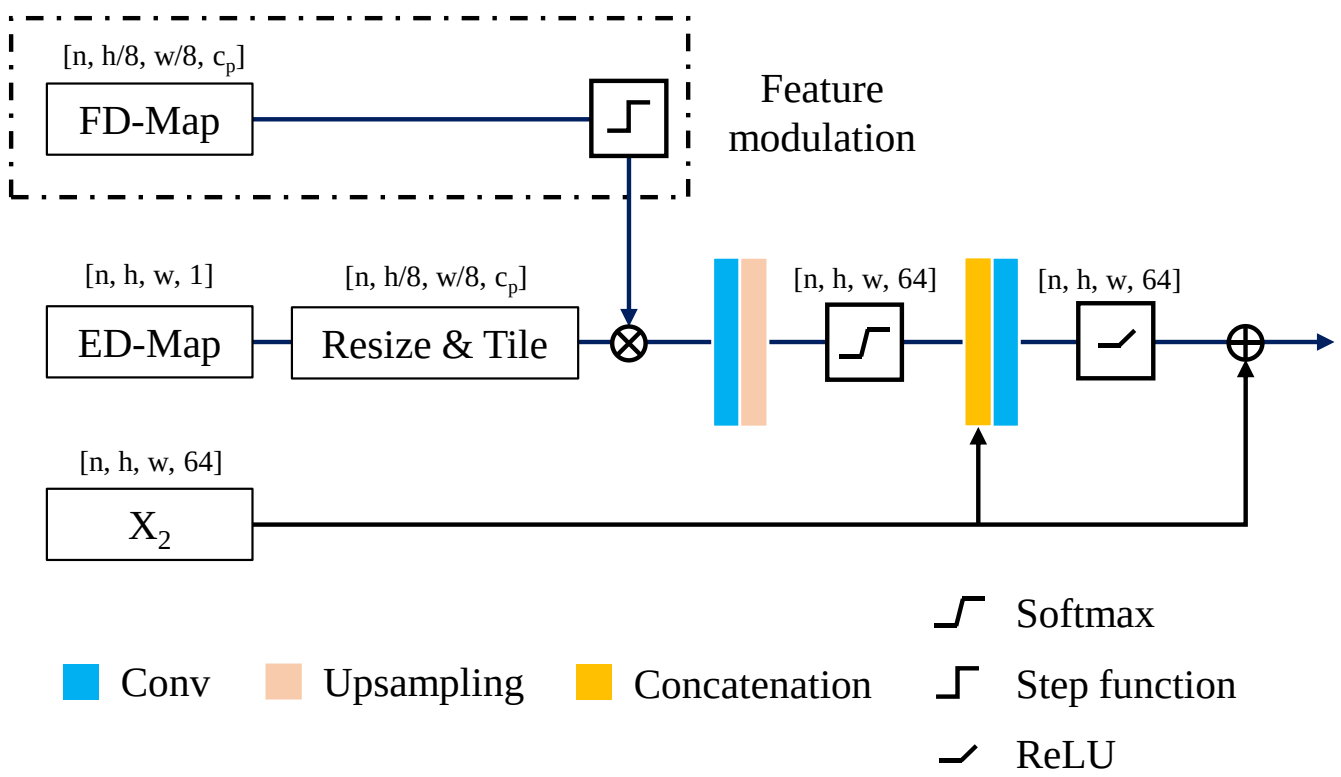

Fig. 4. Edge compensation module. $X_2$ represents the feature map that needs edge compensation in IDAN.

The feature difference attention module (FDA-module) is shown in Fig. 3. On the one hand, the FDA-module combines the features extracted from the original image by the pre-training model to enrich the internal features of IDAN. On the other hand, it obtains the difference results of high-dimensional features so that it can perceive the changing area of the image and optimize the features in IDAN.

The edge compensation module (EC-module) is shown in Fig. 4. The EC-module combines features extracted by the back-bone and the edge information extracted by the Canny operator in ED-map to compensate the edge of the feature map in IDAN. In order to improve the compatibility between single-channel edge features and multi-channel image features, FD-map and step function are used to perform feature modulation on ED-map.

### D. *Difference method and loss function*

The difference operation in Fig. 1 includes feature subtract-tion, expansion and erosion operations:

$$\mathrm{R} = |A - B| \tag{1}$$

$$\mathrm{R} \oplus \mathrm{K} = \{\mathrm{x}, \mathrm{y} \mid (\mathrm{K})_{\mathrm{xy}} \cap \mathrm{R} \neq \emptyset\} \tag{2}$$

$$\mathrm{R} \ominus \mathrm{K} = \{\mathrm{x}, \mathrm{y} \mid (\mathrm{K})_{\mathrm{xy}} \subseteq \mathrm{R}\} \tag{3}$$

A and B represent the feature maps corresponding to the two input images, R represents the result of the two subtractions, and K represents the kernel that performs the expansion or erosion operation.

The Bray-Curtis distance (BCD) is a measure of the dissimilarity between features and has been used [13] as a loss function:

$$\mathrm{BCD} = \frac{\sum_{k=1}^{n} |x_k - y_k|}{\sum_{k=1}^{n} x_k + \sum_{k=1}^{n} y_k} \tag{4}$$

Where, x represents the output of the network prediction, y represents the corresponding label, n represents the total number of pixels in the image, and k represent that our loss needs to be calculated pixel by pixel. If all the values in the input are greater than 0, then the value of BCD will be between 0 and 1. The normalization operation is achieved by dividing the absolute value difference by the sum. IDAN will use BCD as the loss function.

## III. EXPERIMENTAL RESULTS AND ANALYSIS

### A. *Datasets and image preprocessing*

The experiment uses WHU Building change detection dataset (WHU) [14] and LEVIR building Change Detection dataset (LEVIR-CD) [15]. There are two images with a size of 32507×15354 in WHU. These two images were taken in February 2011 and April 2012, covering 12,796 buildings within 20.5 square kilometers. A sliding window with a size of 512×512 is used to sample the original images in sequence, and take one image out of every five images as the test set, and the rest as the training set. A total of 1462 groups of images in the training set and 365 groups of images in the test set were obtained. Each group of images includes images before and after the building changes, and a corresponding label. There are 637 groups of images with a size of 1024×1024 in LEVIR-CD. Each group of images includes the images before and after the building changes, and a corresponding label. The image data collection time span ranges from 5 to 14 years, covering various types of buildings, such as residences, garages, warehouses, etc. This dataset has pre-allocated 445 groups of pictures for training, 64 groups of pictures for verification, and 128 groups of pictures for testing. Each picture was cropped into four 512×512 images for the next step.

Before training the network, we performed a series of preprocessing operations on the image: (1) Zero padding was performed on the original image in four directions, and the size of the image after padding was 1024×1024. (2) Randomly sample the picture using a rectangular frame with a side length in the range of 0.7×512 to 1.3×512. (3) Normalize the sampled image size to 512×512. (4) Perform random rotation and illumination transformation operations within a certain range of the sampling results.

Four metrics were used to evaluate the detection performance of the network, including accuracy (A), precision (P), recall (R), and F1–score (F1):

$$\mathrm{Accuracy} = \frac{\mathrm{TP+TN}}{\mathrm{TP+FP+TN+FN}} \tag{5}$$

$$\mathrm{Precision} = \frac{\mathrm{TP}}{\mathrm{TP+FP}} \tag{6}$$

$$\mathrm{Recall} = \frac{\mathrm{TP}}{\mathrm{TP+FN}} \tag{7}$$

$$\mathrm{F1\text{-}score} = \frac{2\mathrm{TP}}{2\mathrm{TP} + \mathrm{FP} + \mathrm{FN}} \tag{8}$$

TABLE I
EXPERIMENTAL RESULTS OF DIFFERENT IMPROVED METHODS ON WHU

| Methods | Difference map preparation related variables | | | Metrics | | | | GFLOPs | |
|---|---|---|---|---|---|---|---|---|---|
| | Pre-training model (Parameters) | Pre-training model output shape | Edge detection operator | A | P | R | F1 | Backbone | Pre-training model |
| U-net | - | - | - | 99.02% | 90.51% | 85.72% | 88.05% | 160.76 | - |
| U-net + EC-module + FDA-module (IDAN) | MobileNetV2 (65,920) | (N,64,64,192) | Canny | 99.00% | 88.69% | 87.57% | 88.13% | 215.82 | 0.68 |
| | Xception (193,312) | (N,64,64,256) | Canny | 99.09% | **91.57%** | 86.29% | 88.85% | 215.91 | 4.55 |
| | ResNet50V2 (1,171,456) | (N,64,64,512) | Canny | 99.07% | 90.29% | 87.05% | 88.64% | 216.24 | 15.62 |
| | VGG16 (7,635,264) | (N,64,64,512) | Canny | **99.14%** | 90.50% | **88.85%** | **89.67%** | 216.24 | 73.10 |
| | VGG16 (7,635,264) | (N,64,64,512) | Sobel | 99.04% | 90.10% | 86.69% | 88.36% | 216.24 | 73.10 |
| | VGG16 (7,635,264) | (N,64,64,512) | Prewitt | 99.05% | 89.15% | 88.10% | 88.62% | 216.24 | 73.10 |

TABLE II
EXPERIMENTAL RESULTS OF DIFFERENT IMPROVED METHODS ON TWO DATASETS

| Datasets | WHU | | | | LEVIR-CD | | | |
|---|---|---|---|---|---|---|---|---|
| Metrics / Methods | Accuracy | Precision | Recall | F1-score | Accuracy | Precision | Recall | F1-score |
| U-net [5] | 99.02% | 90.51% | 85.72% | 88.05% | 98.84% | 91.98% | 86.37% | 89.09% |
| U-net + SE [6] | 98.94% | 88.61% | 85.96% | 87.27% | 98.86% | 91.36% | 87.48% | 89.14% |
| U-net + Non-local [7] | 98.95% | 88.03% | 87.25% | 87.64% | 98.91% | 91.84% | 88.05% | 89.85% |
| U-net + CBAM [8] | 99.01% | 90.91% | 85.25% | 87.99% | 98.89% | 91.86% | 87.64% | 89.64% |
| U-net + EC-module | 99.01% | 89.64% | 86.54% | 88.06% | 98.96% | 92.25% | 88.31% | 90.24% |
| U-net + FDA-module | 99.13% | **90.94%** | 87.89% | 89.39% | 98.94% | **92.32%** | 87.42% | 90.01% |
| IDAN | **99.14%** | 90.50% | **88.85%** | **89.67%** | **99.04%** | 91.14% | **91.19%** | **91.17%** |

TABLE III
EXPERIMENTAL RESULTS OF DIFFERENT METHODS ON TWO DATASETS

| Datasets | WHU | | | | LEVIR-CD | | | |
|---|---|---|---|---|---|---|---|---|
| Metrics / Methods | Accuracy | Precision | Recall | F1-score | Accuracy | Precision | Recall | F1-score |
| FCN-8s [20] | 98.79% | 90.14% | 80.48% | 85.04% | 98.49% | 87.79% | 84.45% | 86.09% |
| CD-net [21] | 98.91% | 89.42% | 85.48% | 87.41% | 97.83% | 90.73% | 86.82% | 88.73% |
| U-net [5] | 99.02% | 90.51% | 85.72% | 88.05% | 98.86% | **92.93%** | 85.74% | 89.19% |
| U-net++ [11] | 99.05% | **90.92%** | 85.84% | 88.31% | 98.07% | 92.18% | 87.84% | 89.96% |
| DSAMNet [9] | 99.02% | 88.43% | 88.21% | 88.32% | 98.95% | 92.08% | 88.36% | 90.18% |
| DDCNN [10] | 99.07% | 90.90% | 86.43% | 88.61% | 98.11% | 91.85% | 88.69% | 90.24% |
| IDAN | **99.14%** | 90.50% | **88.85%** | **89.67%** | **99.04%** | 91.14% | **91.19%** | **91.17%** |

where, TP is true positive, TF is true negative, FP is false positive, and FN is false negative.

### *B. Experimental result*

There are many optional variables in the preparation stage of the difference map, and the related series of comparative experiment results are shown in table I. For different pre-training models, we uniformly use the feature map after three downsampling operations as the output to match the size of the internal feature map of IDAN. As a representative of the light-weight model, MobileNetV2 [16] improves the detection performance of IDAN with little time cost increase. Xception [17], ResNet50V2 [18] and VGG16 [19] can output higher-dimensional features, which increases the time cost but greatly improves the performance of IDAN. When IDAN is used, we can select an appropriate pre-training model according to actual needs, complete image preprocessing first, and then, release the pre-training model in the GPU to reduce memory usage in the next stage. In subsequent experiments, we all use VGG16 as the pre-training model. Prewitt, Sobel and Canny are all traditional edge detection operators. Canny is a multi-stage optimization operator with filtering, enhancement, and detection. Table I shows the superior performance of Canny operator, so it will be used in subsequent experiments.

In order to verify the optimization effect of FDA-module and EC-module on the model, the experimental results of adding FDA-module and EC-module to the backbone (U-net) separately or simultaneously are shown in table II, and the position of adding the module is shown in Fig. 2. The experiments show that applying the traditional attention mechanism to the remote sensing image change detection task can improve the recall rate to a certain extent, but it will reduce the accuracy rate and lead to a decrease in the F1 score. Both FDA-module and EC-module optimize the model better than the traditional attention mechanism when used alone. With the combined use of FDA-module and EC-module, the F1 scores of the proposed IDAN experiments on WHU and LEVIR-CD improved by 1.62% and 1.98% over the U-net (baseline model), respectively. The experiments demonstrate that both FDA-module and EC-module can optimize the experimental results of the model on the remote sensing image change detection task. Compared with the traditional attention mechanism, the proposed (IDAN) feature optimization module has superior performance. Some of the detection results are shown in Fig. 5.

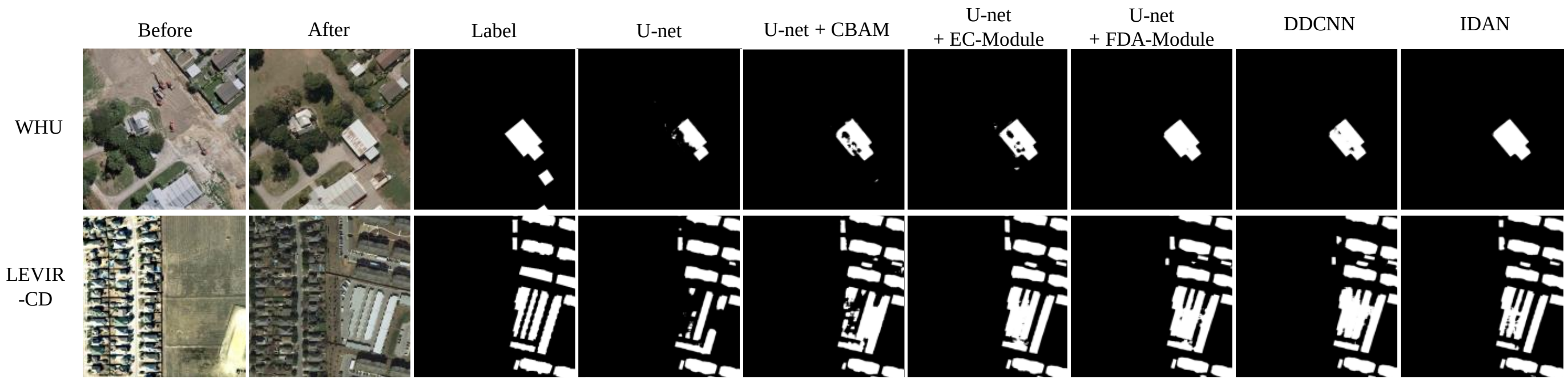

Fig. 5. Change detection results for different methods.

To verify the superiority of IDAN performance, various traditional methods were compared, and the experimental results are shown in table III. Because the attention comes from outside the backbone, IDAN is also affected by some noise while acquiring prior knowledge, which causes a decrease in precision. However, due to the substantial improvement of recall, IDAN still reflects a better overall performance on WHU and LEVIR-CD datasets.

## IV. CONCLUSION

This paper proposes a method for remote sensing image change detection based on image difference attention network (IDAN). The proposed method uses traditional image processing approaches to provide prior knowledge for the network. Through the weighting of regional features and the compensation of feature edges, IDAN shows superior performance in remote sensing image change detection tasks. This research provides a new direction for future research: how to use prior knowledge to guide the model's attention. In the future, the image pre-processing stage can provide more valuable information for model training, such as obtaining richer image features with the help of pre-trained models or using traditional algorithms to provide targeted image information.

## REFERENCES

[1] A. J. Lipton, H. Fujiyoshi and R. S. Patil, "Moving target classification and tracking from real-time video," Proceedings Fourth IEEE Workshop on Applications of Computer Vision. WACV'98 (Cat. No.98EX201), 1998, pp. 8-14, doi: 10.1109/ACV.1998.732851.

[2] F. Zhou, J. Huang, B. Sun, G. Wen and Y. Tian, "Intelligent Identification Method for Natural Disasters along Transmission Lines Based on Inter-FFD-mape Difference and Regional Convolution Neural Network," 2019 IEEE Intl Conf on Parallel & Distributed Processing with Applications, Big Data & Cloud Computing, Sustainable Computing & Communications, Social Computing & Networking (ISPA/BDCloud/SocialCom/SustainCom), 2019, pp. 218-222, doi: 10.1109/ISPA-BDCloud-SustainCom-SocialCom48970.2019.00040.

[3] F. Gao and Y. Lu, "Moving Target Detection Using Inter-FFD-mape Difference Methods Combined with Texture Features and Lab Color Space," 2019 International Conference on Artificial Intelligence and Advanced Manufacturing (AIAM), 2019, pp. 76-81, doi: 10.1109/-AIAM48774.2019.00022.

[4] L. Xu, W. Jing, H. Song and G. Chen, "High-Resolution Remote Sensing Image Change Detection Combined With Pixel-Level and Object-Level," in IEEE Access, vol. 7, pp. 78909-78918, 2019, doi: 10.1109/ACCESS.-2019.2922839.

[5] R. Olaf, P. Fischer, and T. Brox. "U-net: Convolutional networks for biomedical image segmentation." International Conference on Medical image computing and computer-assisted intervention. Springer, Cham, 2015.

[6] J. Hu, L. Shen and G. Sun, "Squeeze-and-Excitation Networks," 2018 IEEE/CVF Conference on Computer Vision and Pattern Recognition, 2018, pp. 7132-7141, doi: 10.1109/CVPR.2018.00745.

[7] X. Wang, R. Girshick, A. Gupta and K. He, "Non-local Neural Networks," 2018 IEEE/CVF Conference on Computer Vision and Pattern Recognition, 2018, pp. 7794-7803, doi: 10.1109/CVPR.2018.00813.

[8] W. Sanghyun, et al. "Cbam: Convolutional block attention module." Proceedings of the European conference on computer vision (ECCV). 2018.

[9] Q. Shi, M. Liu, S. Li, X. Liu, F. Wang and L. Zhang, "A Deeply Supervised Attention Metric-Based Network and an Open Aerial Image Dataset for Remote Sensing Change Detection," in IEEE Transactions on Geoscience and Remote Sensing, doi: 10.1109/TGRS.2021.3085870.

[10] X. Peng, R. Zhong, Z. Li and Q. Li, "Optical Remote Sensing Image Change Detection Based on Attention Mechanism and Image Difference," in IEEE Transactions on Geoscience and Remote Sensing, vol. 59, no. 9, pp. 7296-7307, Sept. 2021, doi: 10.1109/TGRS.2020.3033009.

[11] Z. Zhou, M. M. R. Siddiquee, N. Tajbakhsh, and J. Liang, “Unet++: A nested u-net architecture for medical image segmentation,” in Deep Learn. Med. Image Anal. Multimodal Learn. for Clin. Decis. Support: Springer, vol. 2018, pp. 3–11.

[12] K. Alex, I. Sutskever, and G. E. Hinton. "Imagenet classification with deep convolutional neural networks." Advances in neural information processing systems 25 (2012): 1097-1105.

[13] H. Liu, Q. Ding, Z. Hu and X. Chen, "Remote Sensing Image Vehicle Detection Based on Pre-Training and Random-Initialized Fusion Network," in IEEE Geoscience and Remote Sensing Letters, doi: 10.1109/LGRS.2021.3109637.

[14] S. Ji, S. Wei and M. Lu, "Fully Convolutional Networks for Multisource Building Extraction From an Open Aerial and Satellite Imagery Data Set," in IEEE Transactions on Geoscience and Remote Sensing, vol. 57, no. 1, pp. 574-586, Jan. 2019, doi: 10.1109/TGRS.2018.2858817.

[15] C. Hao, and Z. Shi. "A spatial-temporal attention-based method and a new dataset for remote sensing image change detection." Remote Sensing 12.10 (2020): 1662.

[16] M. Sandler, A. Howard, M. Zhu, A. Zhmoginov and L. Chen, "MobileNetV2: Inverted Residuals and Linear Bottlenecks," 2018 IEEE/CVF Conference on Computer Vision and Pattern Recognition, 2018, pp. 4510-4520, doi: 10.1109/CVPR.2018.00474.

[17] F. Chollet, "Xception: Deep Learning with Depthwise Separable Convolutions," 2017 IEEE Conference on Computer Vision and Pattern Recognition (CVPR), 2017, pp. 1800-1807, doi: 10.1109/CVPR.2017.195.

[18] H. Kaiming, et al. "Identity mappings in deep residual networks." European conference on computer vision. Springer, Cham, 2016.

[19] S. Karen, and A. Zisserman. "Very deep convolutional networks for large-scale image recognition." arXiv preprint arXiv:1409.1556 (2014).

[20] E. Shelhamer, J. Long and T. Darrell, "Fully Convolutional Networks for Semantic Segmentation," in IEEE Transactions on Pattern Analysis and Machine Intelligence, vol. 39, no. 4, pp. 640-651, 1 April 2017, doi: 10.1109/TPAMI.2016.2572683.

[21] P. F. Alcantarilla, S. Stent, G. Ros, R. Arroyo, and R. Gherardi, “Street-view change detection with deconvolutional networks,” Auto. Robots, vol. 42, no. 7, pp. 1301–1322, Oct. 2018.